\documentclass[conference]{IEEEtran}
\IEEEoverridecommandlockouts

\usepackage{cite}
\usepackage{amsmath,amssymb,amsfonts}
\usepackage{booktabs}
\usepackage{algorithm}
\usepackage{algorithmic}
\usepackage{graphicx}
\usepackage{xcolor}
\usepackage{multirow}
\usepackage{array}
\usepackage{eso-pic}
\usepackage[hidelinks]{hyperref}
\usepackage{custom_commands}
\usepackage{stmaryrd}

\graphicspath{{figs/}}

\newtheorem{assumption}{Assumption}
\newtheorem{proposition}{Proposition}

\def\BibTeX{{\rm B\kern-.05em{\sc i\kern-.025em b}\kern-.08em
    T\kern-.1667em\lower.7ex\hbox{E}\kern-.125emX}}

\begin{document}

\title{Large-Scale Bayesian Tensor Reconstruction via Approximate Message Passing}

\author{
\IEEEauthorblockN{Bingyang Cheng, Zhongtao Chen, Yichen Jin, Hao Zhang, Chen Zhang, Edmund Y. Lam\textsuperscript{*}, and Yik-Chung Wu\textsuperscript{*}}
\IEEEauthorblockA{\textit{Department of Electrical and Computer Engineering} \\
\textit{The University of Hong Kong}\\
Hong Kong, China \\
\textsuperscript{*}Corresponding authors: Edmund Y. Lam (\href{mailto:elam@eee.hku.hk}{elam@eee.hku.hk}) and Yik-Chung Wu (\href{mailto:ycwu@eee.hku.hk}{ycwu@eee.hku.hk})}
}

\maketitle

\AddToShipoutPictureFG*{%
  \AtPageLowerLeft{%
    \raisebox{0.38in}{%
      \makebox[\paperwidth][c]{%
        \parbox{7.2in}{\centering\fontsize{6.5}{7.5}\selectfont
          \textcopyright~2026 IEEE. Personal use of this material is permitted. Permission from IEEE must be obtained for all other uses, in any current or future media, including reprinting/republishing this material for advertising or promotional purposes, creating new collective works, for resale or redistribution to servers or lists, or reuse of any copyrighted component of this work in other works.}%
      }%
    }%
  }%
}

\begin{abstract}
While CANDECOMP/PARAFAC (CP) decomposition (CPD) is fundamental for tensor reconstruction, Bayesian CPD often scales poorly because variational updates require repeated matrix inversions.
We develop CP generalized approximate message passing (CP-GAMP) for incomplete noisy Bayesian CPD.
The algorithm uses Gaussian message approximations to avoid high-dimensional inversions, and it combines a Bernoulli-Gaussian prior with expectation-maximization updates to estimate effective CP rank and noise variance.
We also give a formal state evolution (SE) recursion and relate its fixed points to replica-symmetric saddle points, so CP-GAMP's SE-predicted error can be compared with the formal replica-symmetric minimum mean-squared error (MMSE) benchmark in the matched limit.
Synthetic and image-inpainting experiments show that CP-GAMP substantially reduces runtime relative to variational Bayesian CPD while maintaining competitive reconstruction accuracy.
Source code is available at \url{https://github.com/BingyangCheng/CP-GAMP.git}.
\end{abstract}

\begin{IEEEkeywords}
Tensor decomposition, tensor completion, approximate message passing, Bayesian inference, state evolution.
\end{IEEEkeywords}

\section{Introduction}
Tensors provide a natural representation for multiway data in data mining and machine learning, including user-item-time interactions, relational events, spatio-temporal measurements, hyperspectral images, and scientific recordings \cite{anandkumar2014tensor,sidiropoulos2017tensor,williams2018unsupervised,xu2021overfitting}.
In these applications, the goal is often not only to fill in missing entries, but also to discover a compact set of latent factors that explain how different modes interact.
CANDECOMP/PARAFAC (CP) decomposition (CPD) is especially appealing because it expresses a tensor as a sum of rank-one components with factors tied to the original modes \cite{harshman1970foundations,kolda2009tensor}.
This interpretability makes CPD a useful model for exploratory data analysis, recommendation, anomaly detection, and signal reconstruction.

For real data, observations are typically noisy, heavily incomplete, and collected under an unknown effective rank.
Classical optimization methods such as alternating least squares \cite{kolda2009tensor} can be efficient, but they usually require the rank to be chosen externally and do not provide posterior uncertainty.
Bayesian CPD addresses these issues by placing priors on factor matrices and inferring latent factors, noise levels, and rank-related hyperparameters within one probabilistic model \cite{chu2009probabilistic,xiong2010temporal,zhao2015bayesiana}.
The difficulty is computational.
Sampling-based Bayesian tensor methods are often too slow for large tensors \cite{rai2014scalable}, while variational inference (VI)-based Bayesian CPD methods repeatedly update covariance terms and perform matrix inversions whose cost grows rapidly with the rank and the mode dimensions \cite{zhao2015bayesiana,budzinskiy2023variational,fang2022bayesian,tao2023scalable,tao2024efficient}.
As a result, scalable Bayesian inference for large incomplete CP tensors remains challenging.

Approximate message passing (AMP) offers a different route to scalable Bayesian inference.
AMP and generalized AMP (GAMP) replace high-dimensional posterior computations by scalar denoising steps plus residual corrections, yielding efficient algorithms with state evolution characterizations in several high-dimensional models \cite{donoho2009message,rangan2011generalized,javanmard2013state}.
Bilinear GAMP (Bi-GAMP) extends this principle to matrix factorization \cite{parker2014bilineara,parker2014bilinear}, and tensor completion AMP (TC-AMP) applies bilinear inference after unfolding a tensor into matrices \cite{li2016approximate}.
Nevertheless, unfolding can discard the component-sharing structure that distinguishes CPD from independent matrix factorizations.
Existing tensor AMP analyses also tend to focus on idealized tensor estimation models, known priors, or fixed hyperparameters, which limits their direct use in incomplete real-data settings \cite{kadmon2018statistical}.
These limitations motivate an AMP-style Bayesian CPD method that operates directly on the CP factor graph, preserves the multilinear structure, and estimates effective rank and noise parameters.

This paper develops CP generalized approximate message passing (CP-GAMP), a scalable AMP algorithm for Bayesian CP tensor reconstruction from incomplete noisy observations.
Starting from the CPD factor graph, we approximate the high-dimensional loopy belief-propagation messages by Gaussian messages and retain Onsager-style correction terms that compensate for self-feedback in the dense graph.
The resulting updates avoid the matrix inversions used by variational Bayesian CPD and reduce inference to multilinear aggregation, scalar output updates, and scalar factor denoising.
To make the method adaptive, we combine CP-GAMP with a Bernoulli-Gaussian (BG) prior and expectation-maximization (EM) updates \cite{dempster1977maximum,vila2013expectation} so that inactive components can be pruned and the noise variance can be estimated from the data.

The main contributions are as follows.
First, we derive CP-GAMP as a direct AMP approximation on the CP factor graph, rather than applying matrix factorization after tensor unfolding.
Second, we introduce an adaptive CP-GAMP variant that uses a BG prior and EM updates to estimate component activity probabilities and noise variance, supporting data-driven effective-rank selection.
Third, we provide a formal state evolution (SE) recursion for the matched Gaussian-prior large-system limit and relate its informative fixed points to replica-symmetric saddle point equations under explicit assumptions.
Finally, we evaluate CP-GAMP on synthetic tensor reconstruction and image inpainting, including ablations on prior mismatch and non-Gaussian noise.

\section{Related Work}
Classical tensor decomposition methods such as CPD and Tucker decomposition generalize matrix factorization to multiway data \cite{tucker1966some,harshman1970foundations,kolda2009tensor}.
Probabilistic CPD models handle missing entries and uncertainty \cite{chu2009probabilistic,xiong2010temporal}, while Bayesian CP factorization and nonnegative tensor completion can also determine the rank automatically \cite{zhao2015bayesiana,yang2025bayesian}.
Their sampling or variational implementations remain costly at scale because of repeated posterior computations, including covariance updates and matrix inversions.

AMP was originally developed for compressed sensing and high-dimensional linear regression \cite{donoho2009message,javanmard2013state}.
GAMP handles separable nonlinear output channels \cite{rangan2011generalized}, while Bi-GAMP addresses bilinear matrix factorization \cite{parker2014bilineara,parker2014bilinear}.
TC-AMP applies AMP to tensor completion by unfolding a tensor into matrices \cite{li2016approximate}, but this transformation does not directly enforce CP component consistency across all modes.
The CP-GAMP algorithm developed here keeps the CP factor graph and introduces rank and noise learning in the message passing loop.

Statistical physics and information theory have also studied tensor estimation through AMP, state evolution, and replica arguments \cite{montanari2014statistical,lesieur2017statistical,barbier2017layered}.
These works provide asymptotic thresholds and Bayes-optimal predictions for idealized rank-one tensor CPD models.
In contrast, our CP-GAMP construction is written for CPD with an arbitrary working rank $R$, so the message updates and SE discussion retain the multilinear coupling among multiple rank-one components.
The analysis follows the same tradition but is framed carefully: the Gaussian message approximation and the replica-symmetric characterization are modeling assumptions for the large-system analysis.

Nonparametric iterative machine teaching casts example selection as functional optimization \cite{zhang2023nonparametric}, with related methods accelerating implicit neural representations, graph property learners, and attention learners \cite{zhang2026nint,zhang2025nonparametric,zhang2026nonparametric}.
These works reduce neural-network training cost; CP-GAMP instead reduces Bayesian tensor-inference cost on a multilinear factor graph.

\section{Bayesian CPD Model}
For an $N$-th order tensor $\ten{Z}\in\mathbb{R}^{I_1\times\cdots\times I_N}$, the CPD model used in \cite{zhao2015bayesiana} is
\begin{equation}
    \ten{Z}=\sum_{r=1}^{R}\v{a}^{(1)}_{\cdot r}\circ\cdots\circ\v{a}^{(N)}_{\cdot r}
    =\llbracket \m{A}^{(1)},\ldots,\m{A}^{(N)} \rrbracket,
    \label{eq:cpd}
\end{equation}
where $\m{A}^{(n)}\in\mathbb{R}^{I_n\times R}$ is the mode-$n$ factor matrix, $\v{a}^{(n)}_{\cdot r}$ is its $r$-th column, $\circ$ is the vector outer product, and the double brackets denote the CP/Kruskal composition rather than concatenation.
The entry indexed by $\v{i}=(i_1,\ldots,i_N)$ is
\begin{equation}
    z_{\v{i}}=\sum_{r=1}^{R}\prod_{n=1}^{N}a^{(n)}_{i_n,r}.
    \label{eq:entry}
\end{equation}

We observe a noisy and incomplete tensor
\begin{equation}
    \ten{Y}=\ten{O}\odot(\ten{Z}+\ten{W}),
    \label{eq:obs}
\end{equation}
where $\ten{O}$ is a binary observation mask with entries $o_{\v{i}}$, $\Omega=\{\v{i}:o_{\v{i}}=1\}$ is the observed index set, $\odot$ is elementwise multiplication, and $w_{\v{i}}\sim\mathcal{N}(0,v^w)$ is additive white Gaussian noise (AWGN).
For $\v{i}\in\Omega$, $p(y_{\v{i}}\vert z_{\v{i}})=\mathcal{N}(y_{\v{i}};z_{\v{i}},v^w)$; for $\v{i}\notin\Omega$, the likelihood carries no information about $z_{\v{i}}$.

To learn the effective CP rank, we place a BG prior on each factor entry:
\begin{equation}
    p(a^{(n)}_{i_n,r})=(1-\lambda_r)\delta(a^{(n)}_{i_n,r})
    +\lambda_r \mathcal{N}(a^{(n)}_{i_n,r};0,1).
    \label{eq:bg_prior}
\end{equation}
The shared activity parameter $\lambda_r$ controls whether component $r$ is active across all modes.
The posterior is proportional to the separable likelihood times the product of these factor priors:
\begin{equation}
    p(\{\m{A}^{(n)}\}_{n=1}^{N}\vert \ten{Y})
    \propto
    \prod_{\v{i}\in\Omega}p\left(y_{\v{i}}\middle\vert \sum_{r=1}^{R}\prod_{n=1}^{N}a^{(n)}_{i_n,r}\right)
    \prod_{n,i_n,r}p(a^{(n)}_{i_n,r}).
    \label{eq:posterior}
\end{equation}

\section{Method}
\subsection{Factor Graph and Message Approximation}
\begin{figure*}[t]
\centering
\begin{minipage}[t]{0.57\textwidth}
    \centering
    \includegraphics[width=\linewidth]{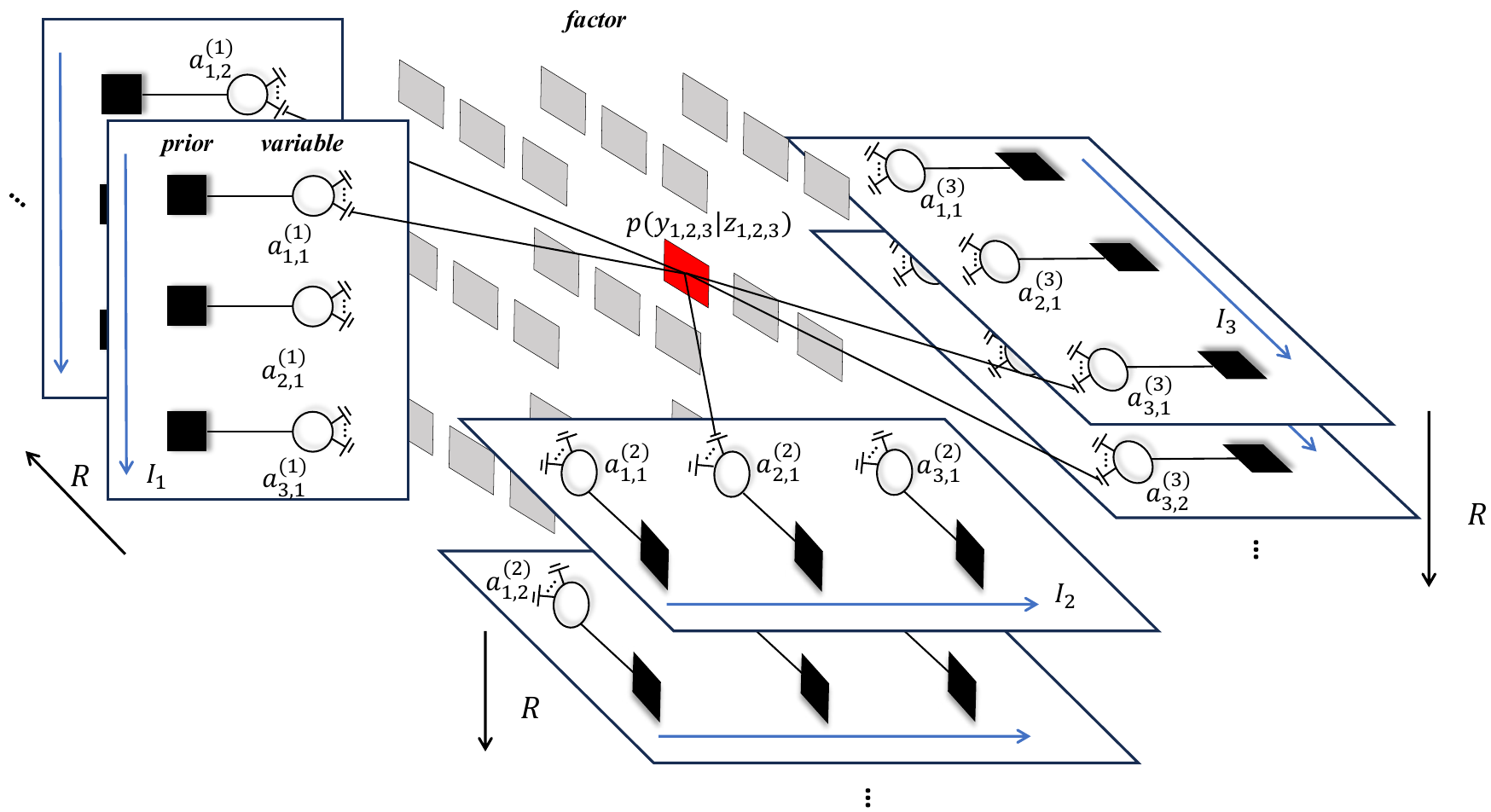}
    {\footnotesize (a) Complete CP factor graph}
\end{minipage}
\hfill
\begin{minipage}[t]{0.38\textwidth}
    \centering
    \includegraphics[width=\linewidth]{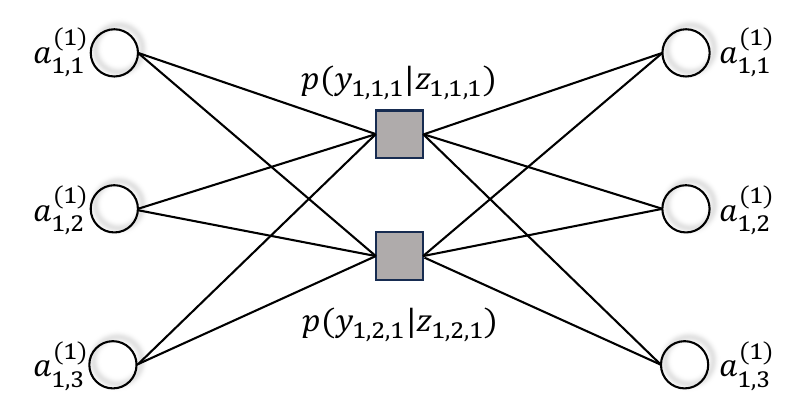}
    {\footnotesize (b) Local short-loop structure}
\end{minipage}
\caption{Factor-graph view of Bayesian CPD. Black squares are prior factors, white circles are factor variables, and gray or red squares are likelihood factors. The plates span mode sizes $I_n$ and component index $R$. Panel (b) isolates two likelihood nodes sharing variables; unshared neighbors are omitted.}
\label{fig:factor_graph}
\end{figure*}

The factor graph contains one likelihood node for each observed tensor entry and one variable node for each factor entry.
Fig.~\ref{fig:factor_graph}(a) illustrates the complete CP factor graph.
The graph is dense and loopy: for instance, two observations that share mode indices are connected to common factor variables, creating short cycles through likelihood nodes and shared variable nodes, as shown in Fig.~\ref{fig:factor_graph}(b).
Exact loopy belief propagation is therefore computationally infeasible because factor-to-variable messages require integrals over all neighboring factor entries \cite{pearl1988probabilistic,kschischang2001factor,frey1997revolution}.

\begin{assumption}[Gaussian message approximation]
\label{assump:gma}
In the large-system regime where $I_1,\ldots,I_N,R$ grow with fixed ratios $I_n/R$, the aggregate messages entering each scalar denoiser can be approximated by Gaussian random variables whose means and variances are tracked by scalar state variables.
\end{assumption}

Assumption~\ref{assump:gma} is the standard AMP-style approximation used to obtain tractable updates and state-evolution characterizations \cite{donoho2009message,rangan2011generalized,javanmard2013state}.
Under this approximation, CP-GAMP reduces the high-dimensional message updates to scalar posterior mean and variance computations.

\subsection{Algorithm}
Let $\hat{a}^{(n)}_{i_n,r}(t)$ and $v^{a^{(n)}}_{i_n,r}(t)$ denote the approximate posterior mean and variance at iteration $t$.
For each tensor entry, CP-GAMP forms the predicted mean and variance
\begin{align}
    \hat{p}_{\v{i}}(t)
    &=\sum_{r=1}^{R}\prod_{n=1}^{N}\hat{a}^{(n)}_{i_n,r}(t), \label{eq:p_update}\\
    v^p_{\v{i}}(t)
    &=\sum_{r=1}^{R}\left[
    \prod_{n}\left(v^{a^{(n)}}_{i_n,r}(t)+(\hat{a}^{(n)}_{i_n,r}(t))^2\right)
    -\prod_{n}(\hat{a}^{(n)}_{i_n,r}(t))^2\right].
    \label{eq:vp_update}
\end{align}
Combining this Gaussian prediction with the likelihood gives the approximate posterior moments $\hat{z}_{\v{i}}(t)$ and $v^z_{\v{i}}(t)$.
For the AWGN missing-data channel,
\begin{align}
\hat{z}_{\v{i}}(t)&=
\begin{cases}
\dfrac{y_{\v{i}}v^p_{\v{i}}(t)+\hat{p}_{\v{i}}(t)v^w}{v^p_{\v{i}}(t)+v^w}, & o_{\v{i}}=1,\\
\hat{p}_{\v{i}}(t), & o_{\v{i}}=0,
\end{cases} \label{eq:z_mean}\\
v^z_{\v{i}}(t)&=
\begin{cases}
\left(1/v^p_{\v{i}}(t)+1/v^w\right)^{-1}, & o_{\v{i}}=1,\\
v^p_{\v{i}}(t), & o_{\v{i}}=0.
\end{cases}
\label{eq:z_var}
\end{align}
The scaled residual and inverse residual variance are
\begin{equation}
    \hat{s}_{\v{i}}(t)=\frac{\hat{z}_{\v{i}}(t)-\hat{p}_{\v{i}}(t)}{v^p_{\v{i}}(t)},
    \quad
    v^s_{\v{i}}(t)=\frac{1}{v^p_{\v{i}}(t)}
    \left(1-\frac{v^z_{\v{i}}(t)}{v^p_{\v{i}}(t)}\right).
    \label{eq:s_update}
\end{equation}

For a factor variable $a^{(n)}_{i_n,r}$, let $\mathcal{F}^{(n)}_{i_n,r}$ denote the set of observed entries adjacent to it, write $\mathcal{F}=\mathcal{F}^{(n)}_{i_n,r}$ in the following equations, and define $\hat{a}^{(\setminus n)}_{\v{i},r}(t)=\prod_{\ell\neq n}\hat{a}^{(\ell)}_{i_\ell,r}(t)$.
The effective scalar observation variance and mean are
\begin{align}
    v^q_{i_n,r}(t)
    &=
    \left(
    \sum_{\v{i}'\in\mathcal{F}}
    \left(\hat{a}^{(\setminus n)}_{\v{i}',r}(t)\right)^2 v^s_{\v{i}'}(t)
    \right)^{-1}, \label{eq:vq_update}\\
    \hat{q}_{i_n,r}(t)
    &=\hat{a}^{(n)}_{i_n,r}(t)
    +v^q_{i_n,r}(t)
    \sum_{\v{i}'\in\mathcal{F}}
    \hat{a}^{(\setminus n)}_{\v{i}',r}(t)\hat{s}_{\v{i}'}(t)
    +\Delta_{\mathrm{Ons}},
    \label{eq:q_update}
\end{align}
where the Onsager correction is
\begin{equation}
    \Delta_{\mathrm{Ons}}
    =
    \hat{a}^{(n)}_{i_n,r}(t)v^q_{i_n,r}(t)
    \sum_{\v{i}'\in\mathcal{F}}
    \omega^{(\setminus n)}_{\v{i}',r}(t)
    \left((\hat{s}_{\v{i}'}(t))^2-v^s_{\v{i}'}(t)\right),
    \label{eq:onsager}
\end{equation}
with
\begin{equation}
    \omega^{(\setminus n)}_{\v{i},r}(t)
    =
    \prod_{\ell\neq n}
    \left(v^{a^{(\ell)}}_{i_\ell,r}(t)+(\hat{a}^{(\ell)}_{i_\ell,r}(t))^2\right)
    -
    \prod_{\ell\neq n}
    (\hat{a}^{(\ell)}_{i_\ell,r}(t))^2.
    \label{eq:omega_excluding}
\end{equation}
This term comes from the first-order variance perturbation retained in the AMP expansion \cite{rangan2011generalized,parker2014bilineara,zou2021multi}.
The denoiser then computes
\begin{align}
    \hat{a}^{(n)}_{i_n,r}(t+1)
    &=g(\hat{q}_{i_n,r}(t),v^q_{i_n,r}(t)), \nonumber\\
    v^{a^{(n)}}_{i_n,r}(t+1)
    &=v^q_{i_n,r}(t)g'(\hat{q}_{i_n,r}(t),v^q_{i_n,r}(t)).
    \label{eq:denoiser}
\end{align}
For the BG prior, $g$ is the posterior mean under a scalar Gaussian observation and has a closed form.
The posterior activity probability is
\begin{equation}
    \pi(\hat{q},v^q;\lambda_r)=
    \left[
    1+
    \left(
    \frac{\lambda_r}{1-\lambda_r}
    \frac{\mathcal{N}(\hat{q};0,1+v^q)}
         {\mathcal{N}(\hat{q};0,v^q)}
    \right)^{-1}
    \right]^{-1}.
    \label{eq:pi}
\end{equation}
Given this activity probability, the BG denoiser is
\begin{align}
    g(\hat{q},v^q)
    &=\pi(\hat{q},v^q;\lambda_r)\gamma(\hat{q},v^q), \label{eq:bg_g}\\
    \gamma(\hat{q},v^q)
    &=\frac{\hat{q}/v^q}{1+1/v^q}, \quad
    \nu(\hat{q},v^q)=\frac{1}{1+1/v^q}. \label{eq:bg_gamma}
\end{align}
The corresponding posterior variance is
\begin{equation}
    v^q g'(\hat{q},v^q)
    =
    \pi(\nu+|\gamma|^2)-\pi^2|\gamma|^2,
    \label{eq:bg_var}
\end{equation}
where $\pi$, $\gamma$, and $\nu$ use the same arguments.
Equations~\eqref{eq:bg_g}-\eqref{eq:bg_var} make the rank-learning mechanism transparent: when the scalar observation for a component is not sufficiently supported by the data, $\pi$ decreases, the posterior mean shrinks toward zero, and EM subsequently reduces $\lambda_r$.

\begin{algorithm}[t]
\caption{Adaptive CP-GAMP}
\label{alg:adaptive}
\begin{algorithmic}[1]
\REQUIRE $\ten{Y}$, mask $\ten{O}$, maximum rank $R$, maximum iterations $T$, pruning threshold $T_\lambda$
\STATE Initialize $\hat{a}^{(n)}_{i_n,r}$, $v^{a^{(n)}}_{i_n,r}$, $\lambda_r$, and $v^w$
\FOR{$t=1$ to $T$}
    \STATE Update $\hat{p}_{\v{i}}$, $v^p_{\v{i}}$ using \eqref{eq:p_update}-\eqref{eq:vp_update}
    \STATE Update $\hat{z}_{\v{i}}$, $v^z_{\v{i}}$, $\hat{s}_{\v{i}}$, $v^s_{\v{i}}$ using \eqref{eq:z_mean}-\eqref{eq:s_update}
    \STATE Update $\hat{q}_{i_n,r}$, $v^q_{i_n,r}$ using \eqref{eq:vq_update}-\eqref{eq:q_update}
    \STATE Denoise each factor entry using \eqref{eq:denoiser}
    \STATE Update $\lambda_r$ and $v^w$ using \eqref{eq:lambda_em}-\eqref{eq:noise_em}
    \STATE Remove component $r$ if $\lambda_r<T_\lambda$
\ENDFOR
\RETURN factor estimates and reconstructed tensor $\hat{\ten{Z}}$
\end{algorithmic}
\end{algorithm}

\subsection{EM Hyperparameter Learning}
The EM update for the BG activity probability is
\begin{equation}
    \lambda_r^{j+1}
    =
    \frac{1}{\sum_{n=1}^{N}I_n}
    \sum_{n=1}^{N}\sum_{i_n=1}^{I_n}
    \pi(\hat{q}_{i_n,r},v^q_{i_n,r};\lambda_r^j).
    \label{eq:lambda_em}
\end{equation}
The noise variance update is
\begin{equation}
    v^{w,j+1}
    =
    \frac{1}{|\Omega|}
    \sum_{\v{i}\in\Omega}
    \left(|y_{\v{i}}-\hat{z}_{\v{i}}|^2+v^z_{\v{i}}\right).
    \label{eq:noise_em}
\end{equation}
The pruning step in Algorithm~\ref{alg:adaptive} is used after the EM update to convert the over-parameterized model into a parsimonious CP representation.
It is a practical model selection step.

\subsection{Computational Complexity and Implementation}
For a dense tensor, the dominant computation in each CP-GAMP iteration is the sweep over tensor entries used to update $\hat{p}_{\v{i}}$, $v^p_{\v{i}}$, $\hat{s}_{\v{i}}$, and $v^s_{\v{i}}$.
For an incomplete tensor, this sweep can be restricted to observed entries, while scalar denoising and BG activity updates scan the factor entries.
The per-iteration complexity is therefore
\begin{equation}
    O\left(NR|\Omega|+R\sum_{n=1}^{N}I_n\right),
\end{equation}
where $\Omega$ is the observed index set.
The first term is the multilinear prediction and residual aggregation over observations, and the second term is the factor-side scalar denoising and activity update cost.
The noise-variance update costs $O(|\Omega|)$ and is lower order.
In contrast, VI-based Bayesian CPD methods perform mode-wise covariance updates whose cost grows cubically in the rank in the matrix-inversion step and also requires repeated Khatri-Rao products \cite{zhao2015bayesiana}.
The advantage of CP-GAMP is most visible when $|\Omega|$ is large but the target rank is moderate or over-specified.

The update equations are separable over tensor entries and over factor rows after the residuals are computed.
This gives a direct parallel implementation: entry-wise updates can be distributed across observed indices, while denoising can be performed independently for every factor entry.

\section{State Evolution and Replica-Symmetric Characterization}
We now summarize the formal asymptotic approximation used to interpret CP-GAMP performance.
The analysis is based on the following assumptions.

\begin{assumption}[Large-system limit]
$I_1,\ldots,I_N,R\to\infty$ with fixed ratios $\alpha_n=I_n/R$.
\end{assumption}

\begin{assumption}[Matched generative model]
Factor entries are generated independently from the prior used by the denoiser, the observation mask is independent and identically distributed (i.i.d.) Bernoulli with probability $\rho$, and the noise is i.i.d. Gaussian with variance $v^w$.
\end{assumption}

Let $\chi_a^{(n)}=\mathbb{E}[(a^{(n)})^2]$ and let $q_a^{(n)}(t)$ denote the estimate second moment tracked by SE.
Define $\chi_z=R\prod_{n=1}^{N}\chi_a^{(n)}$.
Starting from a weakly informative initialization $q_a^{(n)}(0)=\rho_0\chi_a^{(n)}$ with a small $0<\rho_0<1$, the SE recursion is
\begin{align}
    V(t) &= \chi_z-R\prod_{n=1}^{N}q_a^{(n)}(t), \label{eq:se_v_current}\\
    \Sigma_a^{(n)}(t+1)
    &= \frac{v^w+V(t)}
    {\rho\prod_{\ell\neq n}I_\ell q_a^{(\ell)}(t)}, \label{eq:se_sigma}\\
    q_a^{(n)}(t+1)
    &= \mathbb{E}\left[
    \left(\mathbb{E}[a^{(n)}\vert a^{(n)}+\sqrt{\Sigma_a^{(n)}(t+1)}\xi]\right)^2
    \right], \label{eq:se_q}\\
    V(t+1)
    &= \chi_z-R\prod_{n=1}^{N}q_a^{(n)}(t+1),
    \label{eq:se_v}
\end{align}
where $\xi\sim\mathcal{N}(0,1)$.
For a Gaussian prior with variance $\chi_a^{(n)}$, \eqref{eq:se_q} reduces to
\begin{equation}
    q_a^{(n)}(t+1)
    =
    \frac{(\chi_a^{(n)})^2}
    {\chi_a^{(n)}+\Sigma_a^{(n)}(t+1)}.
\end{equation}

\begin{proposition}[Conditional correspondence]
\label{prop:se_replica}
Under the large-system, matched-prior, and replica-symmetry assumptions, and when CP-GAMP converges to the informative fixed point, the formal fixed-point equations of \eqref{eq:se_v_current}-\eqref{eq:se_v} match the replica-symmetric saddle point equations for the corresponding CPD model.
\end{proposition}

A derivation of this SE-replica equivalence, following the SE and replica strategy of multi-layer Bi-GAMP (ML-BiGAMP) \cite{zou2021multi} and the classical replica method \cite{mezard1988spin} but adapted to the CP tensor factor graph, is given in Appendix~\ref{app:replica}.
This also connects the replica calculation to the Bayes benchmark in tensor estimation \cite{lesieur2017statistical,barbier2017layered}: at the replica-symmetric optimum, the tensor-error order parameter $V^\star$ predicts the normalized minimum mean-squared error (MMSE) $V^\star/\chi_z$.
SE tracks the branch reached by CP-GAMP from its initialization. 
It matches this formal MMSE benchmark only when the stable informative fixed point coincides with the replica-symmetric optimum, while coexistence of branches indicates a possible algorithmic gap.
For $N=2$, the CP model reduces to a bilinear factorization of the ML-BiGAMP type, but arbitrary order requires tracking mode-wise overlaps and multilinear Onsager terms, so the lift is not automatic.
This is a formal correspondence rather than a finite-sample convergence result for adaptive BG-EM. 
The SE comparison below uses matched Gaussian-prior CP-GAMP without EM hyperparameter learning or pruning.

\section{Experiments}
\subsection{Experiment Setup}
Synthetic tensors are generated from \eqref{eq:cpd} with factor entries drawn from $\mathcal{N}(0,1)$.
AWGN is added at a target signal-to-noise ratio (SNR), and entries are independently sampled according to the observation ratio.
We report normalized mean-squared error (NMSE),
\begin{equation}
    \mathrm{NMSE}=20\log_{10}\frac{\|\hat{\ten{Z}}-\ten{Z}\|_F}{\|\ten{Z}\|_F}.
\end{equation}
Unless otherwise stated, synthetic experiments use $100\times100\times100$ tensors, SNR $10$ dB, and 10 independent trials.
Adaptive CP-GAMP uses maximum iteration 1000, stopping threshold $3\times10^{-4}$, initial $v^w=1$, initial $\lambda_r=0.5$, and pruning threshold $T_\lambda=0.3$.
FBCP \cite{zhao2015bayesiana} and TC-AMP \cite{li2016approximate} are configured using their published stopping criteria. 
TC-AMP is given the true rank because it does not learn rank automatically.

\subsection{Synthetic Tensor Reconstruction}
\begin{figure*}[t]
\centering
\begin{minipage}[b]{0.28\textwidth}
    \includegraphics[width=\textwidth]{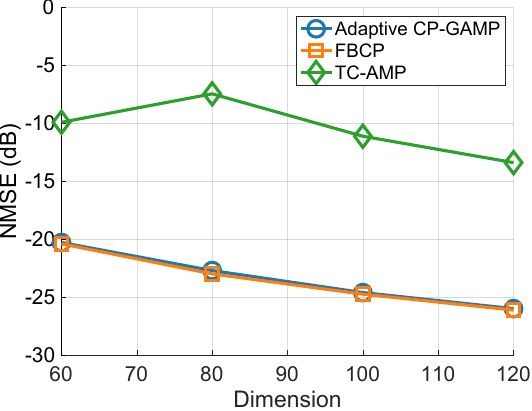}
\end{minipage}
\hfill
\begin{minipage}[b]{0.28\textwidth}
    \includegraphics[width=\textwidth]{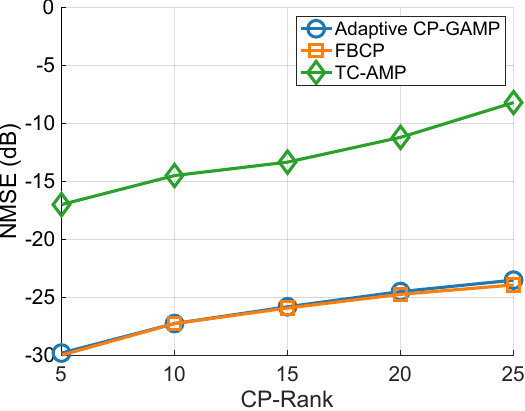}
\end{minipage}
\hfill
\begin{minipage}[b]{0.28\textwidth}
    \includegraphics[width=\textwidth]{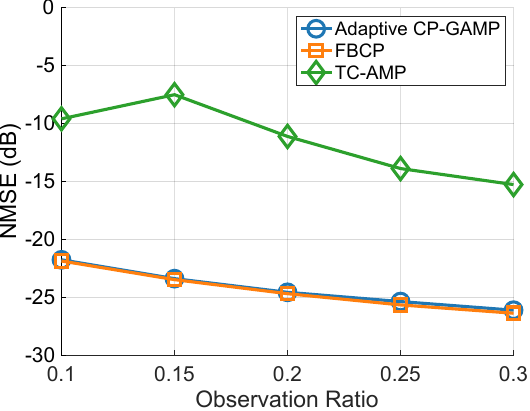}
\end{minipage}
\begin{minipage}[b]{0.28\textwidth}
    \includegraphics[width=\textwidth]{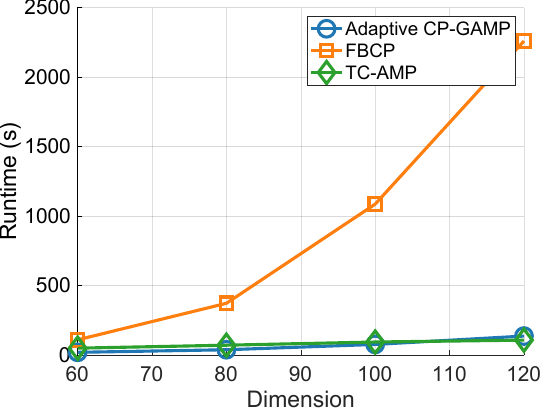}
\end{minipage}
\hfill
\begin{minipage}[b]{0.28\textwidth}
    \includegraphics[width=\textwidth]{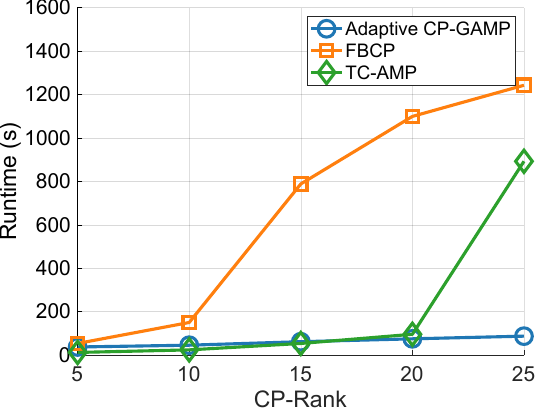}
\end{minipage}
\hfill
\begin{minipage}[b]{0.28\textwidth}
    \includegraphics[width=\textwidth]{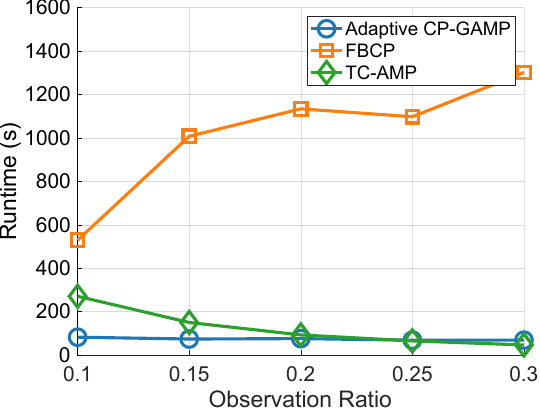}
\end{minipage}
\caption{Synthetic tensor reconstruction at 10 dB SNR. Top: NMSE. Bottom: runtime. Left: varying dimension with rank 20 and observation ratio 0.2. Middle: varying rank for $100^3$ tensors and observation ratio 0.2. Right: varying observation ratio for $100^3$ tensors with rank 20.}
\label{fig:synthetic}
\end{figure*}

Fig.~\ref{fig:synthetic} compares adaptive CP-GAMP with FBCP and TC-AMP.
Adaptive CP-GAMP and FBCP achieve comparable NMSE across tensor dimensions, ranks, and observation ratios. TC-AMP is less accurate in these tests, which is consistent with its unfolded representation not enforcing the shared CP component structure.
The main advantage of CP-GAMP is runtime, since it avoids the matrix inversions required by VI-based updates and scales linearly per iteration in the number of tensor entries.

\subsection{Image Inpainting}
\begin{figure}[t]
\centering
\begin{minipage}[t]{0.18\linewidth}
    \centering
    \includegraphics[width=\linewidth]{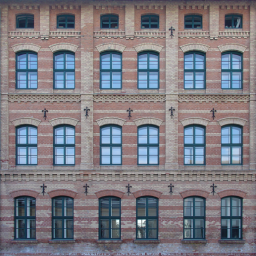}
    {\scriptsize (a) Truth}
\end{minipage}
\begin{minipage}[t]{0.18\linewidth}
    \centering
    \includegraphics[width=\linewidth]{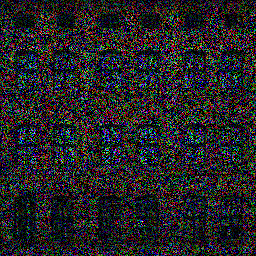}
    {\scriptsize (b) Obs.}
\end{minipage}
\begin{minipage}[t]{0.18\linewidth}
    \centering
    \includegraphics[width=\linewidth]{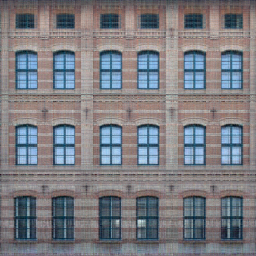}
    {\scriptsize (c) CP-GAMP}
\end{minipage}
\begin{minipage}[t]{0.18\linewidth}
    \centering
    \includegraphics[width=\linewidth]{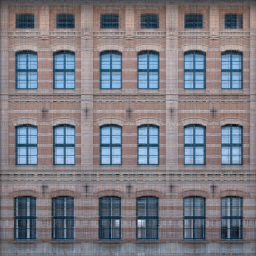}
    {\scriptsize (d) FBCP}
\end{minipage}
\begin{minipage}[t]{0.18\linewidth}
    \centering
    \includegraphics[width=\linewidth]{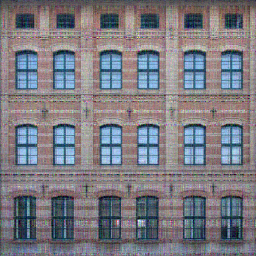}
    {\scriptsize (e) TC-AMP}
\end{minipage}
\caption{Image inpainting with observation ratio 30\% and SNR 10 dB.}
\label{fig:image}
\end{figure}

We also evaluate image inpainting by treating six $256\times256$ images in \cite{zhao2015bayesiana} as third-order tensors.
On the facade image in Fig.~\ref{fig:image}, adaptive CP-GAMP reaches NMSE $-21.43$ dB in 18.99 seconds, compared with $-21.21$ dB in 43.32 seconds for FBCP and $-18.47$ dB in 15.54 seconds for TC-AMP.
In these six image tests, CP-GAMP obtains the lowest NMSE in the table while remaining faster than FBCP.

\begin{table}[t]
\centering
\caption{Image inpainting results at observation ratio 30\% and SNR 10 dB. Runtime is measured in seconds.}
\label{tab:image}
\begin{tabular}{lccc}
\toprule
Image / metric & CP-GAMP & FBCP & TC-AMP \\
\midrule
Airplane NMSE & \textbf{-21.17} & -19.62 & -18.20 \\
Airplane runtime & 24.15 & 64.64 & \textbf{19.20} \\
\midrule
Baboon NMSE & \textbf{-15.13} & -15.08 & -13.35 \\
Baboon runtime & 47.41 & 122.05 & \textbf{19.01} \\
\midrule
Barbara NMSE & \textbf{-16.39} & -15.76 & -14.98 \\
Barbara runtime & \textbf{24.41} & 66.39 & 29.68 \\
\midrule
Facade NMSE & \textbf{-21.43} & -21.21 & -18.47 \\
Facade runtime & 18.99 & 43.32 & \textbf{15.54} \\
\midrule
House NMSE & \textbf{-19.95} & -19.37 & -16.71 \\
House runtime & 22.99 & 96.26 & \textbf{15.95} \\
\midrule
Sailboat NMSE & \textbf{-15.96} & -15.57 & -13.70 \\
Sailboat runtime & \textbf{19.03} & 127.87 & 20.20 \\
\bottomrule
\end{tabular}
\end{table}

Table~\ref{tab:image} shows that CP-GAMP is not always the fastest method on this $256 \times 256$ image set because TC-AMP has a lighter unfolded representation.
However, TC-AMP pays for this speed with lower reconstruction quality in these tests.
FBCP is competitive in NMSE but is slower on all images because its variational updates repeatedly manipulate posterior covariance terms.
The CP-GAMP trade-off is therefore favorable in this image setting when both reconstruction fidelity and runtime matter.

\subsection{State Evolution}
\begin{figure}[t]
\centering
\includegraphics[width=0.78\linewidth]{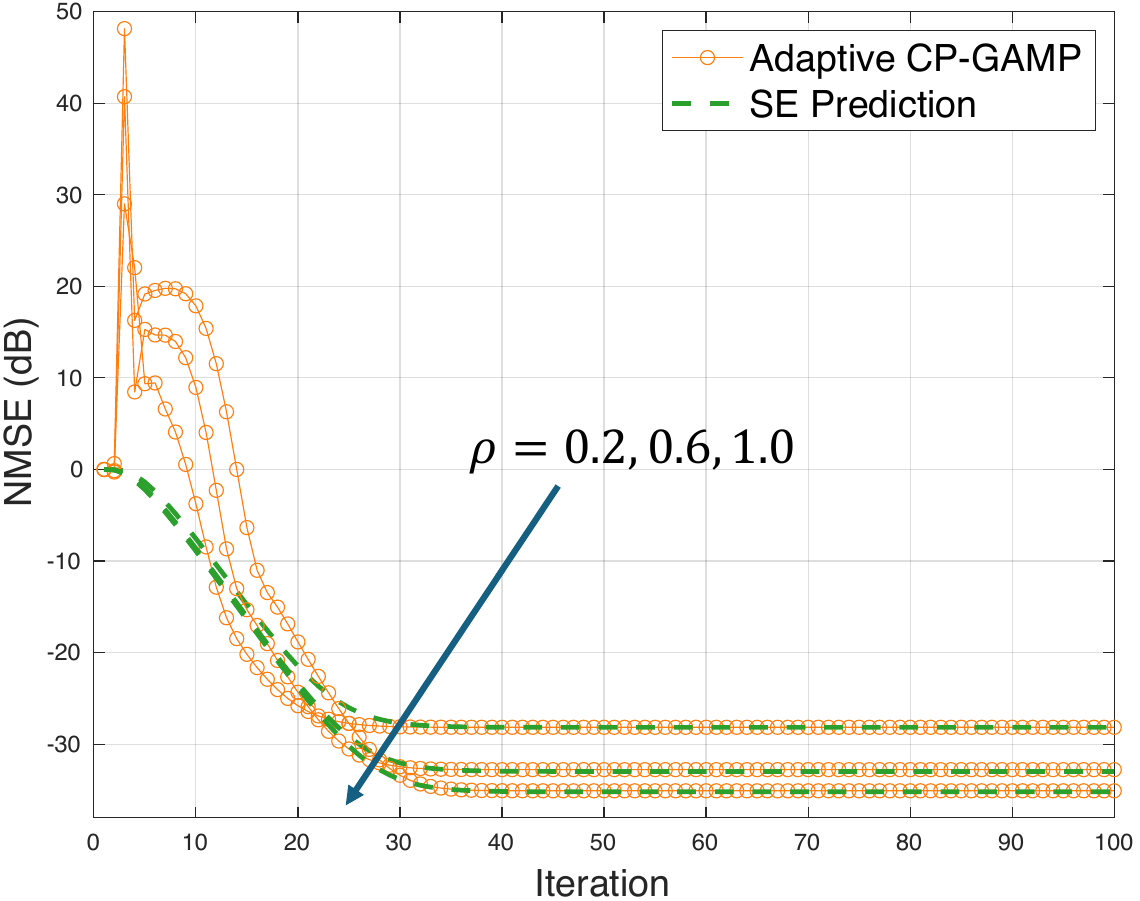}
\caption{SE predictions and empirical CP-GAMP trajectories without EM hyperparameter learning or pruning for $100^3$ rank-20 tensors at 10 dB SNR under observation ratios $\rho\in\{0.2,0.6,1.0\}$.}
\label{fig:se}
\end{figure}

Fig.~\ref{fig:se} compares SE-predicted NMSE curves with empirical CP-GAMP trajectories run without EM hyperparameter learning or pruning.
The curves capture the main convergence trend and the effect of the observation ratio.

\subsection{Ablation and Robustness}
\begin{table}[t]
\centering
\caption{NMSE (dB) versus pruning threshold on $100^3$ tensors with 20\% observed entries.}
\label{tab:pruning}
\begin{tabular}{lccccc}
\toprule
Threshold & R=5 & R=10 & R=15 & R=20 & R=25 \\
\midrule
0.3 & -29.82 & -27.28 & -25.83 & -24.52 & -23.53 \\
0.1 & -29.57 & -26.93 & -25.45 & -24.07 & -22.95 \\
0.0 & -29.50 & -26.91 & -25.41 & -24.04 & -22.91 \\
\bottomrule
\end{tabular}
\end{table}

\begin{table}[t]
\centering
\caption{NMSE (dB) under model mismatch on $100^3$ tensors with 20\% observed entries.}
\label{tab:mismatch}
\begin{tabular}{lccccc}
\toprule
Setting & R=5 & R=10 & R=15 & R=20 & R=25 \\
\midrule
Gaussian var. 1 & -29.82 & -27.28 & -25.83 & -24.52 & -23.53 \\
Gaussian var. 2 & -28.04 & -25.75 & -24.45 & -23.33 & -22.57 \\
Gaussian var. 4 & -26.45 & -24.59 & -23.53 & -22.72 & -22.25 \\
Laplace var. 1 & -27.69 & -27.08 & -25.62 & -22.70 & -22.72 \\
\bottomrule
\end{tabular}
\end{table}

Tables~\ref{tab:pruning} and \ref{tab:mismatch} address robustness to pruning thresholds and model mismatch.
The pruning threshold has only a small effect on reconstruction NMSE over the tested range.
Prior-variance mismatch and variance-matched Laplace noise degrade performance gracefully, usually within a few dB.

\subsection{Discussion}
Across the tested settings, CP-GAMP retains rank and noise learning and factor uncertainty while reducing runtime relative to FBCP.
Its NMSE advantage over TC-AMP is consistent with preserving CP component sharing instead of unfolding the tensor.
The SE curves describe qualitative behavior only: because the comparison fixes rank and hyperparameters, residual gaps may arise from finite size and Gaussian-approximation error.

The implementation evaluates only AWGN, for which \eqref{eq:z_mean}-\eqref{eq:z_var} are closed form.
Other separable channels retain the GAMP interface but require task-specific moments. 
Hence, quadrature or lookup tables add per-entry cost.
Their implementation and real-data validation are left for future work.

\subsection{Limitations}
CP-GAMP assumes low-rank CP structure. 
Strong mode-specific subspaces or non-multilinear structure may require Tucker \cite{tucker1966some}, tensor-train \cite{xu2021overfitting}, or neural models \cite{liu2019costco}.
Its SE-replica analysis is formal, asymptotic, and matched-model, not a finite-sample result for adaptive BG-EM.
Experiments implement AWGN and the BG denoiser. 
Other channels and priors require new scalar moments and, for pruning, suitable hierarchical rules.

\section{Conclusion}
This paper presented CP-GAMP, a scalable Bayesian CP tensor reconstruction algorithm with EM-based rank and noise learning.
The method avoids matrix inversions, keeps the CP multilinear structure, and supports incomplete noisy observations.
We derived a formal SE recursion and related its fixed points to replica-symmetric saddle point equations under explicit large-system and matched-model assumptions.
Experiments show competitive reconstruction accuracy and lower runtime in the tested settings, and ablations suggest robustness to pruning thresholds and moderate model mismatch.

\appendices

\section{Derivation for CP-GAMP}
\label{app:gamp}
This appendix summarizes the derivation that leads from loopy belief propagation to the update equations used in Section~IV \cite{pearl1988probabilistic,kschischang2001factor,frey1997revolution}.
For an observation index $\v{i}=(i_1,\ldots,i_N)$, let
$\mathcal{V}(\v{i})=\{(\ell,k):\ell=1,\ldots,N,\;k=1,\ldots,R\}$ be the set of adjacent scalar factor variables.
The pair $\alpha=(\ell,k)\in\mathcal{V}(\v{i})$ denotes $a_{\v{i},\alpha}=a^{(\ell)}_{i_\ell,k}$.
Let $\Delta^{t}_{\v{i}\to i_n,r}(a^{(n)}_{i_n,r})$ denote the log-message from likelihood node $\v{i}$ to variable node $a^{(n)}_{i_n,r}$, and let $\Delta^{t}_{i_n,r\to\v{i}}(a^{(n)}_{i_n,r})$ denote the reverse message.
For a neighboring variable index $\alpha$, write $m^t_{\alpha\to\v{i}}(a_{\v{i},\alpha})=\exp\{\Delta^t_{\alpha\to\v{i}}(a_{\v{i},\alpha})\}$.
The sum-product updates in log form are
\begin{align}
\Delta^{t}_{\v{i}\to i_n,r}(a)
&=\log\int p(y_{\v{i}}\vert z_{\v{i}})
\prod_{\alpha\in\mathcal{V}(\v{i})\setminus\{(n,r)\}}
m^t_{\alpha\to\v{i}}(a_{\v{i},\alpha})
\nonumber\\
&\qquad\qquad
\,d\mathbf{a}_{\v{i},-(n,r)}+\mathcal{C}, \label{eq:app_f2v}\\
\Delta^{t+1}_{i_n,r\to\v{i}}(a)
&=\log p(a)+
\sum_{\v{i}'\in\mathcal{F}^{(n)}_{i_n,r}\setminus\v{i}}
\Delta^{t}_{\v{i}'\to i_n,r}(a)+\mathcal{C}.
\label{eq:app_v2f}
\end{align}
Here $d\mathbf{a}_{\v{i},-(n,r)}$ integrates over all $a_{\v{i},\alpha}$ with $\alpha\neq(n,r)$, and $\mathcal{C}$ denotes terms independent of the target scalar $a$.
The integral in \eqref{eq:app_f2v} is high-dimensional because $z_{\v{i}}=\sum_{r}\prod_n a^{(n)}_{i_n,r}$ depends on all neighboring factor variables.

For the target component $r$ and mode $n$, we use the notation
$\hat{a}^{(\setminus n)}_{\v{i},r}=\prod_{\ell\neq n}\hat{a}^{(\ell)}_{i_\ell,r}$
for the product of the other-mode means in the same rank-one term.
The corresponding product variance is $\omega^{(\setminus n)}_{\v{i},r}$ in \eqref{eq:omega_excluding}.

Under Assumption~\ref{assump:gma}, the contribution of all variables except $a^{(n)}_{i_n,r}$ is approximated as Gaussian.
Conditioned on $a^{(n)}_{i_n,r}=a$, write
\begin{equation}
    z_{\v{i}}\approx a\,\hat{a}^{(\setminus n)}_{\v{i},r}+\hat{p}^{(\setminus n,r)}_{\v{i}}
    +\epsilon_{\v{i},i_n,r},
    \quad
    \epsilon_{\v{i},i_n,r}\sim\mathcal{N}(0,v^p_{\v{i},i_n,r}).
\end{equation}
Here $\hat{p}^{(\setminus n,r)}_{\v{i}}$ and $v^p_{\v{i},i_n,r}$ collect the mean and variance of the aggregate contribution not represented by the argument $a$ of the target factor entry $a^{(n)}_{i_n,r}$ times $\hat{a}^{(\setminus n)}_{\v{i},r}$.
This reduces the factor-to-variable message to the scalar function
\begin{equation}
    H(\hat{p},v^p;y_{\v{i}})
    =
    \log\int p(y_{\v{i}}\vert z)\mathcal{N}(z;\hat{p},v^p)\,dz.
\end{equation}
Expanding $H$ to second order around the current estimate gives the AMP residual terms
\begin{equation}
    \hat{s}_{\v{i}}=\frac{\partial H}{\partial \hat{p}},
    \quad
    -v^s_{\v{i}}=\frac{\partial^2 H}{\partial \hat{p}^2},
\end{equation}
which are equal to \eqref{eq:s_update} for the Gaussian output channel.
Let $c^{(n)}_{\v{i},r}=\hat{a}^{(\setminus n)}_{\v{i},r}$ and expand the scalar factor message as the target entry is perturbed from its current estimate:
\begin{equation}
    \hat{p}_{\v{i}}(a)
    \approx
    \hat{p}_{\v{i}}+
    (a-\hat{a}^{(n)}_{i_n,r})c^{(n)}_{\v{i},r}.
\end{equation}
The Taylor approximation of the contribution from observation $\v{i}$ is therefore, up to constants independent of $a$,
\begin{equation}
    (a-\hat{a}^{(n)}_{i_n,r})c^{(n)}_{\v{i},r}\hat{s}_{\v{i}}
    -\frac{1}{2}(a-\hat{a}^{(n)}_{i_n,r})^2
    \left(c^{(n)}_{\v{i},r}\right)^2v^s_{\v{i}}.
\end{equation}
Summing these quadratic terms over $\mathcal{F}^{(n)}_{i_n,r}$ gives the edge-dependent Gaussian likelihood
\begin{align}
    \left(\tilde v^q_{i_n,r}\right)^{-1}
    &=
    \sum_{\v{i}\in\mathcal{F}^{(n)}_{i_n,r}}
    \left(c^{(n)}_{\v{i},r}\right)^2v^s_{\v{i}},\\
    \tilde q_{i_n,r}
    &=
    \hat{a}^{(n)}_{i_n,r}
    +\tilde v^q_{i_n,r}
    \sum_{\v{i}\in\mathcal{F}^{(n)}_{i_n,r}}
    c^{(n)}_{\v{i},r}\hat{s}_{\v{i}}.
\end{align}
Equations~\eqref{eq:vq_update}-\eqref{eq:q_update} are the node-aggregated version of these expressions with the first-order Onsager correction retained.
Substituting the quadratic approximation into \eqref{eq:app_v2f} and completing the square yields a Gaussian effective likelihood for each factor variable,
\begin{equation}
    \Delta^{t+1}_{i_n,r}(a)
    \approx
    \log p(a)-\frac{(a-\hat{q}_{i_n,r})^2}{2v^q_{i_n,r}}+\mathcal{C}.
\end{equation}
Taking the posterior mean and variance under this scalar density gives the denoising step \eqref{eq:denoiser}.
The Onsager correction in \eqref{eq:q_update} appears from removing the self-interaction between the current variable and the likelihood nodes that sent messages to it.
This correction distinguishes CP-GAMP from a naive Gaussian loopy belief-propagation (BP) approximation and provides the AMP-style self-feedback cancellation used by the formal SE derivation.

The expressions used in the main text are obtained after replacing edge-dependent quantities by node-dependent ones.
For example, an edge-wise Gaussian approximation would assign a different edge-dependent value $\hat{a}^{(n)}_{i_n,r\to\v{i}}$ to every edge between a factor variable and an observation node.
Keeping all such edge quantities would require memory proportional to $R|\Omega|N$, which is impractical for large tensors.
The AMP expansion suggests that, in the large-system scaling, the difference between an edge quantity and its node aggregate is a lower-order correction once the Onsager term is retained.
Thus CP-GAMP stores only $\hat{a}^{(n)}_{i_n,r}$ and $v^{a^{(n)}}_{i_n,r}$ for every factor entry and uses the residual correction to compensate for the removed self-feedback.

For the AWGN output channel, the derivatives of $H$ can be evaluated without numerical integration.
Combining the Gaussian pseudo-prior $z_{\v{i}}\sim\mathcal{N}(\hat{p}_{\v{i}},v^p_{\v{i}})$ with $y_{\v{i}}\vert z_{\v{i}}\sim\mathcal{N}(z_{\v{i}},v^w)$ gives
\begin{equation}
    p(z_{\v{i}}\vert y_{\v{i}},\hat{p}_{\v{i}},v^p_{\v{i}})
    =
    \mathcal{N}\left(
    z_{\v{i}};
    \frac{y_{\v{i}}v^p_{\v{i}}+\hat{p}_{\v{i}}v^w}{v^p_{\v{i}}+v^w},
    \frac{v^p_{\v{i}}v^w}{v^p_{\v{i}}+v^w}
    \right).
\end{equation}
For a missing entry, the likelihood is uninformative and the posterior equals the pseudo-prior.
These two cases yield \eqref{eq:z_mean}-\eqref{eq:z_var}.
Other separable channels can be used by replacing this scalar posterior computation while leaving the factor-side denoising unchanged.

\section{EM Updates}
\label{app:em}
The EM updates in \eqref{eq:lambda_em}-\eqref{eq:noise_em} follow from maximizing the expected complete-data log-likelihood \cite{dempster1977maximum}.
For the activity probability $\lambda_r$, the relevant complete-data term is
\begin{equation}
    \sum_{n=1}^{N}\sum_{i_n=1}^{I_n}
    \log\left((1-\lambda_r)\delta(a^{(n)}_{i_n,r})
    +\lambda_r\mathcal{N}(a^{(n)}_{i_n,r};0,1)\right).
\end{equation}
Let $b^{(n)}_{i_n,r}$ be the latent Bernoulli indicator that equals one when $a^{(n)}_{i_n,r}$ is active.
The scalar posterior in CP-GAMP gives
\begin{equation}
    \mathbb{E}[b^{(n)}_{i_n,r}\vert\ten{Y}]
    =
    \pi(\hat{q}_{i_n,r},v^q_{i_n,r};\lambda_r^j).
\end{equation}
The expected log-likelihood for $\lambda_r$ is therefore
\begin{align}
    Q(\lambda_r)
    &=
    \sum_{n,i_n}\Big[
    \pi(\hat{q}_{i_n,r},v^q_{i_n,r};\lambda_r^j)\log\lambda_r
    \nonumber\\
    &\qquad\quad
    +\left(1-\pi(\hat{q}_{i_n,r},v^q_{i_n,r};\lambda_r^j)\right)\log(1-\lambda_r)
    \Big]+\mathcal{C}.
\end{align}
Setting $\partial Q/\partial\lambda_r=0$ yields \eqref{eq:lambda_em}.

For the noise variance, the complete-data term is
\begin{equation}
    -\frac{1}{2}\sum_{\v{i}\in\Omega}
    \left[\log v^w+\frac{|y_{\v{i}}-z_{\v{i}}|^2}{v^w}\right].
\end{equation}
The posterior expectation of the squared residual is
\begin{equation}
    \mathbb{E}\left[|y_{\v{i}}-z_{\v{i}}|^2\vert\ten{Y}\right]
    =
    |y_{\v{i}}-\hat{z}_{\v{i}}|^2+v^z_{\v{i}}.
\end{equation}
Maximizing with respect to $v^w$ gives the average residual energy over observed entries, which is exactly \eqref{eq:noise_em}.

\section{Rank Learning and Pruning}
\label{app:rank}
The BG prior introduces a component-level activity probability $\lambda_r$ shared by the $r$-th columns of all factor matrices.
This differs from placing independent sparsity parameters on every factor entry.
The shared parameter encourages entire rank-one components to be removed, which is the desired behavior for CP-rank learning.
At an EM iteration, $\lambda_r$ is updated by averaging scalar posterior activity probabilities over all entries in component $r$:
\begin{equation}
    \lambda_r^{j+1}
    =
    \frac{1}{\sum_n I_n}
    \sum_n\sum_{i_n}\pi(\hat{q}_{i_n,r},v^q_{i_n,r};\lambda_r^j).
\end{equation}
Thus a component remains active only if it is consistently supported across modes and factor rows.
This averaging makes the rank estimate less sensitive to isolated noisy entries than a rule based on individual coefficient magnitudes.

The pruning rule $\lambda_r<T_\lambda$ is not required to define the posterior approximation.
It is used to reduce computation after EM has already assigned low posterior activity to a component.
Without pruning, inactive components remain in the factor matrices with means close to zero but still contribute to the cost of computing \eqref{eq:p_update}-\eqref{eq:vp_update}.
Pruning therefore acts as a computational shortcut and as a way to report an explicit effective rank.
The ablation in Table~\ref{tab:pruning} shows that the reconstruction NMSE changes little when the threshold is reduced from 0.3 to 0.1 or removed entirely, suggesting that the posterior shrinkage is doing most of the statistical work and pruning mainly affects model compactness.

There is also a complementary view of rank learning based on rank identifiability and moment matching in \cite{da2025identifiability}.
Such approaches can provide an initial estimate or upper bound for the rank before message passing starts.
In CP-GAMP, this information could be used to choose the maximum rank $R$ or to initialize $\lambda_r$ non-uniformly.
The BG-EM mechanism would then refine the rank estimate during inference.
This combination may reduce the amount of over-parameterization needed in large tensors and is a useful direction for future work.

\section{SE-Replica Correspondence}
\label{app:replica}
This appendix gives the derivation behind Proposition~\ref{prop:se_replica}.
The argument follows the empirical-convergence and replica-symmetric strategy used for ML-BiGAMP \cite{zou2021multi}, with the bilinear product replaced by the CP multilinear product. 
For $N=2$ this collapses to the ML-BiGAMP-type bilinear case, but the extension to arbitrary order is nontrivial because the mode-wise overlaps and Onsager corrections must be tracked separately.

\noindent\textit{SE derivation.}
Assume empirical convergence with bounded second moments in the pseudo-Lipschitz sense used in AMP SE \cite{javanmard2013state}.
For fixed mode $n$, the CP-GAMP input to $a^{(n)}_{i_n,r}$ is a sum over $O(\prod_{\ell\neq n}I_\ell)$ neighboring entries. 
After the Onsager correction in \eqref{eq:q_update}, the large-system central limit theorem (CLT) motivates the scalar channel
\begin{align}
    \hat q^{(n)}
    &=a^{(n)}+\sqrt{\Sigma_a^{(n)}}\xi, \quad \xi\sim\mathcal{N}(0,1),\\
    q_a^{(n)}
    &=
    \mathbb{E}\!\left[
    \left(\mathbb{E}[a^{(n)}\vert \hat q^{(n)}]\right)^2
    \right].
\end{align}
Under the matched prior, the Nishimori identity \cite{nishimori2001statistical} makes this second moment equal to the planted-estimate overlap.
The CP product then gives
\begin{equation}
    V=\mathbb{E}[(z_{\v{i}}-\hat z_{\v{i}})^2]
    =
    \chi_z-R\prod_{n=1}^{N}q_a^{(n)}.
\end{equation}
To see this product form, write the contribution of component $r$ as
$x_{\v{i},r}=\prod_n a^{(n)}_{i_n,r}$ and its estimate as
$\hat x_{\v{i},r}=\prod_n \hat a^{(n)}_{i_n,r}$.
The zero-mean independent components make the cross terms
$\mathbb{E}[x_{\v{i},r}x_{\v{i},r'}]$ and
$\mathbb{E}[x_{\v{i},r}\hat x_{\v{i},r'}]$ vanish for $r\neq r'$.
For a fixed component, empirical convergence gives
\begin{equation}
    \mathbb{E}[x_{\v{i},r}\hat x_{\v{i},r}]
    =
    \prod_{n=1}^{N}
    \mathbb{E}[a^{(n)}\hat a^{(n)}]
    =
    \prod_{n=1}^{N}q_a^{(n)},
\end{equation}
where the last equality uses the matched-prior Nishimori relation.
Thus the tensor-level overlap is the product of the mode-wise overlaps, multiplied by the number of active rank components.
For the AWGN missing-data channel, $v^p_{\v{i}}\to V$ and averaging over the Bernoulli mask yields
\begin{align}
    v^z
    &=
    \rho\frac{v^wV}{v^w+V}+(1-\rho)V,\\
    v^s
    &=
    \frac{1}{V}\left(1-\frac{v^z}{V}\right)
    =
    \frac{\rho}{v^w+V},\\
    \frac{1}{\Sigma_a^{(n)}}
    &=
    \frac{\rho\prod_{\ell\neq n}I_\ell q_a^{(\ell)}}{v^w+V}.
\end{align}
These identities are equivalent to \eqref{eq:se_v_current}-\eqref{eq:se_v}.

\noindent\textit{Replica correspondence.}
The corresponding formal replica calculation \cite{mezard1988spin} starts from the quenched log partition function, normalized according to the chosen large-system scaling, and applies the identity
\begin{equation}
    \mathbb{E}\log p(\ten{Y})
    =
    \lim_{\tau\to0}\partial_\tau \log \mathbb{E}p^\tau(\ten{Y}).
\end{equation}
For integer $\tau$, the replicated factors share the same observation tensor.
Introduce overlaps
\begin{equation}
    Q_{ab}^{(n)}
    =
    \frac{1}{I_nR}\sum_{i_n,r}
    a_{i_n,r}^{(n)(a)}a_{i_n,r}^{(n)(b)},
    \quad a,b=0,\ldots,\tau,
\end{equation}
where $a=0$ denotes the planted replica.
The CP product gives replicated tensor covariance $R\prod_nQ_{ab}^{(n)}$.
Under the replica-symmetric (RS) ansatz, $Q_{aa}^{(n)}=\chi_a^{(n)}$ and $Q_{ab}^{(n)}=q_a^{(n)}$ for $a\neq b$; the conjugate saddle-point variables decouple each factor entry into the same scalar AWGN channel as SE.
Stationarity gives
\begin{align}
q_a^{(n)}
&=
\mathbb{E}\left[
\left(\mathbb{E}[a^{(n)}\vert a^{(n)}+\sqrt{\Sigma_a^{(n)}}\xi]\right)^2
\right],\\
\Sigma_a^{(n)}
&=
\frac{v^w+V}
{\rho\prod_{\ell\neq n}I_\ell q_a^{(\ell)}},\\
V&=\chi_z-R\prod_{n=1}^{N}q_a^{(n)}.
\end{align}
These match the fixed-point equations of \eqref{eq:se_v_current}-\eqref{eq:se_v}.
This is the fixed-point correspondence summarized in Proposition~\ref{prop:se_replica}.
The same $V$ is the SE per-entry tensor error, $V=\mathbb{E}[(z_{\v{i}}-\hat z_{\v{i}})^2]$.
At the replica-symmetric Bayes branch it gives the formal MMSE prediction, with normalized error $V/\chi_z$.
Thus the SE curve is an algorithmic trajectory toward one saddle-point branch, while the lowest-error Bayes branch gives the formal MMSE benchmark.
The argument relies on replica symmetry, matched priors, and convergence to the informative branch.

The finite-size gap in the no-EM SE comparison may arise because the factor graph has many short cycles and the Gaussian approximation relies on averaging over many weakly dependent contributions.

\section{Experimental Settings and Reproducibility}
\label{app:exp_settings}
In our experiments, we use random initialization unless otherwise stated.
The pruning threshold $T_\lambda=0.3$ is applied only after EM has updated $\lambda_r$, which reduces unnecessary components and accelerates later iterations.

For synthetic experiments, a ground-truth tensor is generated by drawing all factor entries independently from $\mathcal{N}(0,1)$ and then forming the CP tensor in \eqref{eq:cpd}.
The noise variance is selected to match the target SNR,
\begin{equation}
    \mathrm{SNR}=10\log_{10}\frac{\mathrm{Var}(\ten{Z})}{v^w}.
\end{equation}
The observation mask is sampled independently with probability $\rho$ for each entry.
All reported synthetic NMSE values are averages over 10 independent tensors and masks.

The compared methods are configured as follows.
For synthetic experiments, both adaptive CP-GAMP and FBCP use the initial working rank
$R_{\mathrm{init}}=\max(\lfloor \max_n I_n/2\rfloor,R_{\mathrm{true}})$,
where $R_{\mathrm{true}}$ is the ground-truth CP rank.
CP-GAMP then learns the active components through $\lambda_r$.
FBCP is run with automatic rank determination, following the published stopping rule.
TC-AMP is given the true rank because it does not have the same EM rank-learning mechanism.
This gives TC-AMP extra rank information and makes the comparison conservative with respect to rank selection.
The maximum iteration count is 1000 for all iterative methods.
The stopping threshold for CP-GAMP is $3\times10^{-4}$ on the relative change of the reconstructed tensor.

For image inpainting, each red-green-blue (RGB) image is represented as a $256\times256\times3$ tensor.
The missing mask is applied independently to tensor entries, and AWGN is added at 10 dB SNR before masking.
Runtime is measured wall-clock time for the full reconstruction, including hyperparameter updates.
Synthetic runs used an Intel i5-8500 desktop and image runs used an Apple M2 Pro laptop, both with 16 GB memory.
Because hardware differs, runtime comparisons should be interpreted within each experimental group rather than across synthetic and image tasks.

\section{Observation Channels and Priors}
\label{app:channels}
CP-GAMP separates multilinear aggregation from scalar inference.
The output module requires the posterior mean and variance of $z_{\v{i}}$ under a Gaussian pseudo-prior and the derivatives summarized by $\hat{s}_{\v{i}}$ and $v^s_{\v{i}}$.
These are closed form for AWGN.
Unimplemented alternatives include Bernoulli, count, and heavy-tailed likelihoods \cite{rangan2011generalized}.
Their moments require closed forms, lookup tables, or quadrature, with $K$-point quadrature adding $O(K|\Omega|)$ work per iteration.

For any separable prior $p_0(a)$, the factor module uses the scalar channel $\hat{q}=a+w$, $w\sim\mathcal{N}(0,v^q)$, and returns its posterior mean and variance.
Gaussian, BG, and some mixture priors are closed form; nonconjugate priors require numerical evaluation.
The shared BG activity $\lambda_r$ supplies rank shrinkage, whereas other priors need an analogous hierarchy for pruning, and their SE characterization must be rederived.

\section{Complexity Comparison}
\label{app:complexity}
This appendix gives a more explicit comparison between CP-GAMP and the main baselines.
Let $N$ be the tensor order, $R$ the working rank, and $\Omega$ the observed entry set.
For CP-GAMP, the dominant cost per iteration comes from computing the multilinear predictions and residual statistics over $\Omega$, followed by the scalar denoising updates over all factor entries.
This gives the practical cost
\begin{equation}
    O\!\left(NR|\Omega| + R\sum_{n=1}^{N}I_n\right),
\end{equation}
where the first term dominates in large incomplete tensors.
The memory footprint is similarly modest because the algorithm stores scalar means and variances for factor entries and scalar residual terms for observations.

For FBCP \cite{zhao2015bayesiana}, the cost structure is different.
In addition to expectation updates over observed entries, VI requires covariance updates for latent factors and repeated products involving Khatri-Rao structure.
Using the same notation, its per-iteration cost is
\begin{equation}
O\!\left(
NR^2|\Omega|+R^3\sum_{n=1}^{N}I_n
\right).
\end{equation}
These terms reflect rank-dependent sufficient-statistic accumulation, dense covariance updates, and matrix inversions.
Even when efficient implementations reduce constant factors, the resulting runtime grows much faster with $R$ than the scalar-denoiser structure of CP-GAMP.
This explains the widening runtime gap in Fig.~\ref{fig:synthetic} as the rank or tensor dimension increases.

TC-AMP \cite{li2016approximate} is closer to CP-GAMP in spirit because it also avoids VI-style covariance updates.
Its main limitation is not computational but structural.
The unfolding step transforms the tensor problem into a matrix factorization problem and therefore weakens the shared-component interpretation of CP factors across all modes.
This is why TC-AMP can remain fast on the $256 \times 256$ image inpainting tasks while showing lower NMSE accuracy than direct CP-GAMP updates in our tests.

\IEEEtriggeratref{18}
\bibliographystyle{IEEEtran}
\bibliography{icdm_ref}

\begin{thebibliography}{10}
\providecommand{\url}[1]{#1}
\csname url@samestyle\endcsname
\providecommand{\newblock}{\relax}
\providecommand{\bibinfo}[2]{#2}
\providecommand{\BIBentrySTDinterwordspacing}{\spaceskip=0pt\relax}
\providecommand{\BIBentryALTinterwordstretchfactor}{4}
\providecommand{\BIBentryALTinterwordspacing}{\spaceskip=\fontdimen2\font plus
\BIBentryALTinterwordstretchfactor\fontdimen3\font minus \fontdimen4\font\relax}
\providecommand{\BIBforeignlanguage}[2]{{%
\expandafter\ifx\csname l@#1\endcsname\relax
\typeout{** WARNING: IEEEtran.bst: No hyphenation pattern has been}%
\typeout{** loaded for the language `#1'. Using the pattern for}%
\typeout{** the default language instead.}%
\else
\language=\csname l@#1\endcsname
\fi
#2}}
\providecommand{\BIBdecl}{\relax}
\BIBdecl

\bibitem{anandkumar2014tensor}
A.~Anandkumar, R.~Ge, D.~Hsu, S.~M. Kakade, and M.~Telgarsky, ``Tensor decompositions for learning latent variable models,'' \emph{J. Mach. Learn. Res.}, vol.~15, pp. 2773--2832, 2014.

\bibitem{sidiropoulos2017tensor}
N.~D. Sidiropoulos, L.~De~Lathauwer, X.~Fu, K.~Huang, E.~E. Papalexakis, and C.~Faloutsos, ``Tensor decomposition for signal processing and machine learning,'' \emph{{IEEE} Trans. Signal Process.}, vol.~65, no.~13, pp. 3551--3582, 2017.

\bibitem{williams2018unsupervised}
A.~H. Williams, T.~H. Kim, F.~Wang, S.~Vyas, S.~I. Ryu, K.~V. Shenoy, M.~Schnitzer, T.~G. Kolda, and S.~Ganguli, ``Unsupervised discovery of demixed, low-dimensional neural dynamics across multiple timescales through tensor component analysis,'' \emph{Neuron}, vol.~98, no.~6, pp. 1099--1115.e8, 2018.

\bibitem{xu2021overfitting}
L.~Xu, L.~Cheng, N.~Wong, and Y.-C. Wu, ``Overfitting avoidance in tensor train factorization and completion: Prior analysis and inference,'' in \emph{Proc. {IEEE} Int. Conf. Data Mining ({ICDM})}, 2021, pp. 1439--1444.

\bibitem{harshman1970foundations}
R.~A. Harshman, ``Foundations of the {PARAFAC} procedure: Models and conditions for an {``explanatory''} multi-modal factor analysis,'' \emph{{UCLA} Work. Papers in Phonetics}, vol.~16, pp. 1--84, 1970.

\bibitem{kolda2009tensor}
T.~G. Kolda and B.~W. Bader, ``Tensor decompositions and applications,'' \emph{{SIAM} Rev.}, vol.~51, no.~3, pp. 455--500, 2009.

\bibitem{chu2009probabilistic}
W.~Chu and Z.~Ghahramani, ``Probabilistic models for incomplete multi-dimensional arrays,'' in \emph{Proc. Int. Conf. Artif. Intell. Statist.}, ser. Proc. Mach. Learn. Res., vol.~5.\hskip 1em plus 0.5em minus 0.4em\relax PMLR, 2009, pp. 89--96.

\bibitem{xiong2010temporal}
L.~Xiong, X.~Chen, T.-K. Huang, J.~Schneider, and J.~G. Carbonell, ``Temporal collaborative filtering with {Bayesian} probabilistic tensor factorization,'' in \emph{Proc. {SIAM} Int. Conf. Data Mining}, 2010, pp. 211--222.

\bibitem{zhao2015bayesiana}
Q.~Zhao, L.~Zhang, and A.~Cichocki, ``Bayesian {CP} factorization of incomplete tensors with automatic rank determination,'' \emph{{IEEE} Trans. Pattern Anal. Mach. Intell.}, vol.~37, no.~9, pp. 1751--1763, 2015.

\bibitem{rai2014scalable}
P.~Rai, Y.~Wang, S.~Guo, G.~Chen, D.~Dunson, and L.~Carin, ``Scalable {Bayesian} low-rank decomposition of incomplete multiway tensors,'' in \emph{Proc. Int. Conf. Mach. Learn.}, ser. Proc. Mach. Learn. Res., vol.~32, no.~2.\hskip 1em plus 0.5em minus 0.4em\relax PMLR, 2014, pp. 1800--1808.

\bibitem{budzinskiy2023variational}
S.~Budzinskiy and N.~Zamarashkin, ``Variational {Bayesian} inference for {CP} tensor completion with subspace information,'' \emph{Lobachevskii J. Math.}, vol.~44, no.~8, pp. 3016--3027, 2023.

\bibitem{fang2022bayesian}
S.~Fang, A.~Narayan, R.~Kirby, and S.~Zhe, ``{Bayesian} continuous-time {Tucker} decomposition,'' in \emph{Proc. Int. Conf. Mach. Learn.}, ser. Proc. Mach. Learn. Res., vol. 162.\hskip 1em plus 0.5em minus 0.4em\relax PMLR, 2022, pp. 6235--6245.

\bibitem{tao2023scalable}
Z.~Tao, T.~Tanaka, and Q.~Zhao, ``Scalable {Bayesian} tensor ring factorization for multiway data analysis,'' in \emph{Proc. Int. Conf. Neural Inf. Process.}, ser. Lecture Notes in Computer Science, vol. 14447.\hskip 1em plus 0.5em minus 0.4em\relax Singapore: Springer Nature Singapore, 2024, pp. 490--503.

\bibitem{tao2024efficient}
------, ``Efficient nonparametric tensor decomposition for binary and count data,'' in \emph{Proc. {AAAI} Conf. Artif. Intell.}, vol.~38, no.~14, 2024, pp. 15\,319--15\,327.

\bibitem{donoho2009message}
D.~L. Donoho, A.~Maleki, and A.~Montanari, ``Message-passing algorithms for compressed sensing,'' \emph{Proc. Natl. Acad. Sci.}, vol. 106, no.~45, pp. 18\,914--18\,919, 2009.

\bibitem{rangan2011generalized}
S.~Rangan, ``Generalized approximate message passing for estimation with random linear mixing,'' in \emph{Proc. {IEEE} Int. Symp. Inf. Theory}, 2011, pp. 2168--2172.

\bibitem{javanmard2013state}
A.~Javanmard and A.~Montanari, ``State evolution for general approximate message passing algorithms, with applications to spatial coupling,'' \emph{Inf. Inference, J. {IMA}}, vol.~2, no.~2, pp. 115--144, 2013.

\bibitem{parker2014bilineara}
J.~T. Parker, P.~Schniter, and V.~Cevher, ``Bilinear generalized approximate message passing—part {I}: Derivation,'' \emph{{IEEE} Trans. Signal Process.}, vol.~62, no.~22, pp. 5839--5853, 2014.

\bibitem{parker2014bilinear}
------, ``Bilinear generalized approximate message passing—part {II}: Applications,'' \emph{{IEEE} Trans. Signal Process.}, vol.~62, no.~22, pp. 5854--5867, 2014.

\bibitem{li2016approximate}
Y.~Li, C.~Yin, and Z.~Han, ``An approximate message passing approach for tensor-based seismic data interpolation with randomly missing traces,'' in \emph{Proc. {IEEE} Int. Conf. Acoust., Speech Signal Process. ({ICASSP})}, 2016, pp. 1402--1406.

\bibitem{kadmon2018statistical}
J.~Kadmon and S.~Ganguli, ``Statistical mechanics of low-rank tensor decomposition,'' in \emph{Adv. Neural Inf. Process. Syst.}, vol.~31, 2018, pp. 8201--8212.

\bibitem{dempster1977maximum}
A.~P. Dempster, N.~M. Laird, and D.~B. Rubin, ``Maximum likelihood from incomplete data via the {EM} algorithm,'' \emph{J. Roy. Stat. Soc. Ser. B Methodol.}, vol.~39, no.~1, pp. 1--22, 1977.

\bibitem{vila2013expectation}
J.~P. Vila and P.~Schniter, ``Expectation-maximization {Gaussian}-mixture approximate message passing,'' \emph{{IEEE} Trans. Signal Process.}, vol.~61, no.~19, pp. 4658--4672, 2013.

\bibitem{tucker1966some}
L.~R. Tucker, ``Some mathematical notes on three-mode factor analysis,'' \emph{Psychometrika}, vol.~31, no.~3, pp. 279--311, 1966.

\bibitem{yang2025bayesian}
Z.~Yang, L.~T. Yang, H.~Wang, H.~Zhao, and D.~Liu, ``Bayesian nonnegative tensor completion with automatic rank determination,'' \emph{{IEEE} Trans. Image Process.}, vol.~34, pp. 2036--2051, 2025.

\bibitem{montanari2014statistical}
A.~Montanari and E.~Richard, ``A statistical model for tensor {PCA},'' in \emph{Adv. Neural Inf. Process. Syst.}, vol.~27, 2014, pp. 2897--2905.

\bibitem{lesieur2017statistical}
T.~Lesieur, L.~Miolane, M.~Lelarge, F.~Krzakala, and L.~Zdeborov{\'a}, ``Statistical and computational phase transitions in spiked tensor estimation,'' in \emph{Proc. {IEEE} Int. Symp. Inf. Theory ({ISIT})}, 2017, pp. 511--515.

\bibitem{barbier2017layered}
J.~Barbier, N.~Macris, and L.~Miolane, ``The layered structure of tensor estimation and its mutual information,'' in \emph{Proc. 55th Annu. Allerton Conf. Commun., Control, Comput.}, 2017, pp. 1056--1063.

\bibitem{zhang2023nonparametric}
C.~Zhang, X.~Cao, W.~Liu, I.~Tsang, and J.~Kwok, ``Nonparametric iterative machine teaching,'' in \emph{Proc. 40th Int. Conf. Mach. Learn.}, ser. Proc. Mach. Learn. Res., vol. 202.\hskip 1em plus 0.5em minus 0.4em\relax PMLR, 2023, pp. 40\,851--40\,870.

\bibitem{zhang2026nint}
C.~Zhang, W.~Zuo, B.~Cheng, Y.~Wang, W.-B. Kou, Y.-C. Wu, and N.~Wong, ``{NTK}-guided implicit neural teaching,'' in \emph{Proc. {IEEE/CVF} Conf. Comput. Vis. Pattern Recognit.}, 2026.

\bibitem{zhang2025nonparametric}
C.~Zhang, W.~Bu, Z.~Ren, Z.~Liu, Y.-C. Wu, and N.~Wong, ``Nonparametric teaching for graph property learners,'' in \emph{Proc. 42nd Int. Conf. Mach. Learn.}, ser. Proc. Mach. Learn. Res., vol. 267.\hskip 1em plus 0.5em minus 0.4em\relax PMLR, 2025, pp. 74\,515--74\,539.

\bibitem{zhang2026nonparametric}
C.~Zhang, J.~Wang, B.~Cheng, Z.~Chen, W.~Xu, C.~Wang, M.~Canini, F.~Orabona, Y.-C. Wu, and N.~Wong, ``Nonparametric teaching of attention learners,'' in \emph{Proc. Int. Conf. Learn. Represent.}, 2026.

\bibitem{pearl1988probabilistic}
J.~Pearl, \emph{Probabilistic Reasoning in Intelligent Systems: Networks of Plausible Inference}.\hskip 1em plus 0.5em minus 0.4em\relax Morgan Kaufmann, 1988.

\bibitem{kschischang2001factor}
F.~R. Kschischang, B.~J. Frey, and H.-A. Loeliger, ``Factor graphs and the sum-product algorithm,'' \emph{{IEEE} Trans. Inf. Theory}, vol.~47, no.~2, pp. 498--519, 2001.

\bibitem{frey1997revolution}
B.~J. Frey and D.~J.~C. MacKay, ``A revolution: Belief propagation in graphs with cycles,'' in \emph{Adv. Neural Inf. Process. Syst.}, vol.~10, 1997, pp. 479--485.

\bibitem{zou2021multi}
Q.~Zou, H.~Zhang, and H.~Yang, ``Multi-layer bilinear generalized approximate message passing,'' \emph{{IEEE} Trans. Signal Process.}, vol.~69, pp. 4529--4543, 2021.

\bibitem{mezard1988spin}
M.~M{\'e}zard, G.~Parisi, and M.~A. Virasoro, \emph{Spin Glass Theory and Beyond: An Introduction to the Replica Method and Its Applications}, ser. World Scientific Lecture Notes in Physics.\hskip 1em plus 0.5em minus 0.4em\relax Singapore: World Scientific, 1987, vol.~9.

\bibitem{liu2019costco}
H.~Liu, Y.~Li, M.~Tsang, and Y.~Liu, ``{CoSTCo}: A neural tensor completion model for sparse tensors,'' in \emph{Proc. 25th {ACM} {SIGKDD} Int. Conf. Knowl. Discovery Data Mining}, 2019, pp. 324--334.

\bibitem{da2025identifiability}
E.~da~Silva, A.~Klami, D.~Mesquita, and I.~Urteaga, ``On the identifiability of tensor ranks via prior predictive matching,'' 2025, arXiv:2510.14523.

\bibitem{nishimori2001statistical}
H.~Nishimori, \emph{Statistical Physics of Spin Glasses and Information Processing: An Introduction}, ser. International Series of Monographs on Physics.\hskip 1em plus 0.5em minus 0.4em\relax Oxford, U.K.: Oxford University Press, 2001, vol. 111.

\end{thebibliography}

\end{document}